\documentclass[letterpaper, 10 pt, conference]{ieeeconf}  

\IEEEoverridecommandlockouts                              

\usepackage{amsmath}
\usepackage{amssymb}
\usepackage{algorithm}
\usepackage[noend]{algpseudocode}
\usepackage{xcolor}
\usepackage{cite}
\usepackage{graphicx}
\usepackage{booktabs}
\usepackage{pifont}
\usepackage{multirow}
\usepackage{array}

\newtheorem{thm}{Theorem}[section]

\newtheorem{defn}{Definition}[section]

\DeclareMathOperator{\Argmax}{Argmax}
\DeclareMathOperator{\enab}{Enab}

\DeclareMathOperator{\dom}{Dom}

\DeclareMathOperator{\Reach}{Reach}
\title{\LARGE \bf
 Predictive Safety Shield for Dyna-Q Reinforcement Learning
}

\author{Pin Jin$^{1},$ Hanna Krasowski$^{2}$ and Elena Vanneaux$^{1}$
\thanks{*This work was supported in part by the project TRAITS, under the French National Research Agency (ANR) grant number ANR-21-FAI1-0005}
\thanks{$^{1}$ Departement U2IS, ENSTA Paris, Institut Polytechnique de Paris
        {\tt\small pin.jin|elena.vanneaux@ensta-paris.fr}}%
\thanks{$^{2}$ Department of Electrical Engineering and Computer Science, University of California, Berkeley, USA
        {\tt\small krasowski@berkeley.edu}}%
}

\usepackage{tcolorbox}
\usepackage{xcolor}

\tcbset{
    confidential/.style={
        colback=red!5!white, 
        colframe=red!75!black, 
        fonttitle=\bfseries,
        title=CONFIDENTIAL,
        coltitle=white,
        colbacktitle=red!75!black,
        center title,
        boxrule=1mm, 
        enhanced,
    }
}

\begin{document}

\maketitle
\thispagestyle{empty}
\pagestyle{empty}

\begin{abstract}
Obtaining safety guarantees for reinforcement learning is a major challenge to achieve applicability for real-world tasks. Safety shields extend standard reinforcement learning and achieve hard safety guarantees. However, existing safety shields commonly use random sampling of safe actions or a fixed fallback controller, therefore disregarding future performance implications of different safe actions.
In this work, we propose a predictive safety shield for model-based reinforcement learning agents in discrete space.
Our safety shield updates the \mbox{Q-function} locally based on safe predictions, which originate from a safe simulation of the environment model. This shielding approach improves performance while maintaining hard safety guarantees.
Our experiments on gridworld environments demonstrate that even short prediction horizons can be sufficient to identify the optimal path. We observe that our approach is robust to distribution shifts, e.g., between simulation and reality, without requiring additional training. 
\end{abstract}


\section{Introduction}
A common approach to make reinforcement learning (RL) applicable for safety-critical applications is provably safe RL, which extends standard RL with verification methods so that hard safety guarantees are provided \cite{krasowski2023provably}. Provably safe RL can be categorized by the action correction mechanism; Action masking calculates the set of safe actions in a given state and restricts the action selection process to this set. Action replacement and action projection correct an action proposed by the agent if necessary where replacement uses a safe action as correction and projection identifies the closest safe action with respect to some metric. Post-posed approaches, i.e., projection and replacement, are often more computationally efficient since only a safe action has to be found. Additionally, action replacement is often the easiest to implement and has been identified as best performing in a recent benchmarking \cite{krasowski2023provably}. 

The most common action replacement mechanisms are using a fallback controller or sampling from the action space until a safe action is found. However, these mechanisms do not incorporate that some safe actions might be better suited with respect to fulfilling the goal than others. Notably, Alshiekh et al.~\cite{Alshiekh2018} address this by letting the agent propose multiple actions in a ranked order. However, this still requires defining the number of actions to propose apriori. 

An important assumption for provably safe RL is that the safety of an action in a given state can be evaluated \cite{krasowski2023provably}. This usually requires a safety-relevant system model that conservatively describes the behavior of the system. While provably safe RL can be used with model-free and model-based RL agents, model-based approaches can potentially benefit from this model \cite{He2022,goodall2023approximate}. For discrete spaces, \mbox{Dyna-Q} \cite{Sutton1990} is a simple and effective model-based RL algorithm. However, if the environment slightly changes, e.g., a static obstacle is shifted, \mbox{Dyna-Q} requires many training steps to update the model and often follows a suboptimal policy until the model is updated \cite{Sutton1998}.

In this work, we propose a provably safe shield to verify model-based RL agents online. We introduce a local Q-function that includes the effect of the safety shield for multiple future time steps, which leads to a more performant action selection, especially under distribution shifts, which could originate from a sim-to-real gap of the environment model. Our main contributions are:
\begin{itemize}
    \item We integrate a multi-step safety shield into the model-based RL algorithm Dyna-Q; 
    \item We derive the assumptions necessary to ensure optimality of our approach;
    \item We compare our approach to state-of-the-art baselines to empirically investigate the prediction parameters of our safety shield.
\end{itemize}

The remainder of the paper is structured as follows: we discuss related literature in Sec.~\ref{sec:relatedwork} and present preliminary concepts and notation in Sec.~\ref{sec:preliminaries} and Sec.~\ref{sec:provablysafeRL}. In Sec.~\ref{sec:method}, we detail our predictive safety shield approach. Our numerical experiments are described in Sec.~\ref{sec:experiments}. We discuss our findings in Sec.~\ref{sec:discussion} and conclude in Sec.~\ref{sec:conclusion}.






\section{Related Work}\label{sec:relatedwork}
There are two main paradigms for obtaining safety guarantees for RL policies: verifying the policy for all possible states \cite{Landers2023-RLverifysurvey} or safeguarding the policy online \cite{krasowski2023provably}, for both concepts, either probabilistic or hard guarantees are obtained. If the environment is dynamic, often only probabilistic guarantees can be obtained efficiently, e.g., through scenario optimization \cite{Campi2018, Krasowski2023a.probablisticSafeRL}, when verifying the policy for all states. 
Since we aim for hard guarantees, the safeguarding concept is better suited since it can efficiently achieve hard guarantees in dynamic environments as well \cite{krasowski2023provably, Alshiekh2018}. Our approach is based on action replacement, where a post-posed safety verification identifies unsafe actions and replaces them with safe actions. Common verification methods for action verification are model checking \cite{Alshiekh2018, Konighofer2020}, theorem proving 
\cite{Hunt2021}, reachability analysis \cite{Selim2022, Shao2021, Fisac2019} while the applicability depends on the type of state and action space. In this work, we assume that a verification method is provided and focus on the integration of this safety verification with model-based RL for discrete action spaces.

Additionally, let us discuss the existing literature on safeguards for RL in discrete spaces in more detail, which can be split into safeguards obtaining probabilistic guarantees \cite{He2022,Konighofer2020, Hasanbeig2019, Hasanbeig2020,goodall2023approximate, Jansen2020} and hard guarantees \cite{Konighofer2020, Alshiekh2018, Hunt2021, Carr2023}. For instance, He et al.~\cite{He2022} propose a probabilistic shield that samples trajectories to determine the risk probability and interfere if the predicted probability is above a defined threshold. To produce the trajectory, the current policy is iteratively executed in a latent space, i.e., with a learned environment model. Goodall et al.~\cite{goodall2023approximate} generalize this work and derive probability bounds for constraint violation with learned and true environment models. They perform evaluations on five Atari games and show that their advancements lead to higher performance and fewer violations. 
The studies \cite{Alshiekh2018, Konighofer2020} propose safeguarding mechanisms for action replacement and action masking in discrete spaces by synthesizing safety shields, which provide hard guarantees, from linear temporal logic specifications. Hunt et al.~\cite{Hunt2021} identify safe actions through theorem proofing of differential dynamic logic. However, these action replacement approaches \cite{Konighofer2020, Alshiekh2018, Hunt2021} do not consider optimality criteria for selecting a replacement.
In this work, we use an environment model to produce multiple trajectories with an active safeguard that prevents all constraint violations. We use the trajectories to locally update a Q-function so that the agent selects more optimal actions that are likely safe as well.

\section{Preliminaries} \label{sec:preliminaries}

Reinforcement learning commonly utilizes the MDP framework to model the interaction between a learning agent and its environment \cite{Sutton1998}.
\begin{defn}
\label{MDP}
A Markov decision process (MDP) is a tuple $\mathcal{P}=(S, A, T, r),$ where
\begin{itemize}
\item $S$ is a set of states called the state space;
\item $A(s)$ is a set of actions available in a state $s$.  
\item $T(s,a,s') = Pr(s_{t+1}=s'\mid s_t = s, a_t = a)$ is a probability that action $a$ in a state $s$ at time $t$ will lead to state $s'$ at time $t+1.$ One calls $T$ a state transition function;
\item $r(s, a,s')$ is the immediate reward (or expected immediate reward) received after transitioning from state $s$ to state $s'$ due to action $a$.
\end{itemize}
\end{defn}
This paper assumes that the state space $S$ is finite. It is also assumed that sets of actions $A(s)$ are finite and non-empty for every given state $s$. Let us also emphasize here that if a transition $(s,a,s')$ never occurs in the MDP, i.e., it is impossible by
definition or in the structure of the environment, we will assume that $(s,a,s')\notin T,$ i.e., $T(s,a,s')$ is not defined. That is different from the situation when $T(s,a,s') = 0$. The latter means that a transition may occur even if it is highly unlikely.

An RL agent interacts with the environment in discrete time. At every time step $t$, the agent observes the state $s_t$ of the environment. It then chooses an action $a_t$ w.r.t a current policy $\pi$. The environment then moves to a state $s_{t+1}$ with probability $T(s_{t},a_t,s_{t+1})$ and gets the reward $r_t = r(s_{t},a_t,s_{t+1})$. Formally, a policy $\pi$ is a mapping from states of the environment to actions the agent will perform. The agent's objective is to learn the optimal policy $\pi^*$ that maximizes $\max_{\pi\in\Pi}\mathbb{E}_{\pi}R$ over the class of policies $\Pi$. Where $R = \sum_{t=0}^\infty\gamma^tr_t$ is the cumulative future reward $R$, discounted by factor $\gamma \in [0,1)$. There are two main types of policies in RL. A deterministic policy directly maps each state to a single action. For a stochastic policy, actions are chosen based on a probability distribution.
    
For a given state $s$ and action $a$ the optimal action-value function is defined as  $Q^{*}(s, a) = \max_{\pi\in\Pi}Q_{\pi}(s,a),$ where $$Q_{\pi}(s, a)=\mathbb{E}_\pi\left[\sum_{k=0}^\infty \gamma^kr_{t+k+1}\mid s_t = s, a_t= a\right].$$ Note that the theory of MDPs states that a deterministic greedy policy $\pi^*(s) = a^*, a^* \in \Argmax_{a\in A(s)} Q^{*}(s, a)$ is an optimal one \cite{Sutton1998}. 

Reinforcement learning (RL) algorithms can be model-free or model-based, depending on whether they use a model of the environment. Model-free methods learn directly from experience, while model-based methods build a model to plan and make decisions. 
Model-based methods also differ in distribution and sample models. Distribution methods produce a description of all possibilities and their probabilities;
sample models generate reward $r$ and next state $s'$ when given a current state $s$ and action $a$. The samples might be from a simulation or just taken from the history of what the learning algorithm has experienced so far.
While distribution models are more general, sample models are often easier to obtain in practice.

In this paper, we will use two well-known RL algorithms.
First, Q-learning is a model-free, off-policy RL algorithm that aims to learn the quality of actions. At each time step, the agent updates the Q-values using the rule: 
\begin{multline}
\label{QlearningUpdate}
Q(s_t, a_t) \leftarrow  Q(s_t, a_t) \\
             + \alpha \left[ r_t + \gamma \max_{a'} Q(s_{t+1}, a') - Q(s_t, a_t) \right]
\end{multline}
where $\alpha$ is the learning rate and $\gamma$ is the discount factor. Second, Dyna-Q is an extension of Q-learning that integrates model-based and model-free learning, using a sample-based environment model to simulate additional experiences. This hybrid approach allows the agent to learn more efficiently by leveraging real and simulated experiences. We refer the interested reader to \cite{Sutton1998} for more details. 

\section{Provably Safe Reinforcement Learning} \label{sec:provablysafeRL}

RL is a powerful approach for decision-making in robotics, but for safety-critical applications, it's essential to ensure that RL-driven robots avoid unsafe actions. The vulnerability of standard RL-based controllers to failures has spurred significant growth in research on safe RL in the past decade. Existing approaches include modification of the optimization criteria, the introduction of risk sensitivity,
robust model estimation, and risk-directed policy exploration. They may provide probabilistic, soft or hard safety guarantees. We will focus on provably safe RL approaches, ensuring that unsafe actions are never executed~\cite{krasowski2023provably}.

\subsection{Safety-relevant model and fallback controller}

In provably safe RL, safety is verified with an abstract model containing the safety-relevant behavior of the original MDP.

\begin{defn}
\label{SafetyRelevantAbstraction}
For a MDP $\mathcal{P}=(S, A, T, r)$ we define a safety-relevant model $\Sigma_{\mathcal{P}} = (S, U_{A}, F),$ as a transition system  with the same set of states $S$, with the set of available inputs $U_{A}$, such that $\bigcup_{s\in S} A(s)\subseteq U_A$, and with the transition relation $F\subseteq S\times U_A\times S$, such that $(s,a,s')\in F$ iff $(s,a,s') \in T$.
\end{defn}

Let $F(s, a) = \{s'\in S\mid (s,a,s')\in F\}$ be a set of all possible successors of a state $s$, when action $a$ is taken and let $\enab_F(s) = \{a\in U_{A}\mid F(s, a)\neq\emptyset\}$ be a set of all enabled inputs at a state $s$. Let us then define an operator $\Reach_F(s,1) = \cup_{a\in A(s)} F(s,a)$, which consists of all states that are reachable from $s$ in one step. This operator can be naturally extended towards the sets  $\Reach_F(S,1) = \cup_{s'\in S} \Reach_F(s',1)$. We then inductively define the set of all states reachable from $s$ in N steps $\Reach_F(s, N) =\Reach_F(\Reach_F(s, N-1),1)$. And the set $\overline{\Reach}_F(s, N) = \cup_{i=1:N} F(s,i)$ of all states that are reachable from a state $s$ in no more than N steps.

Let a safety specification be given by a sequence of safe sets $ S^S_t\subseteq S, t = 0,1,2\ldots $ that an agent should not leave while moving in the environment. At every time step $t$, a controller for a transition system $\Sigma $ is a map $C_t\colon S \rightarrow 2^{U_A} $, such that $C_t(s)\subseteq \enab_F(s)$ for every $s\in S.$
\begin{defn}
\label{SafetyController}
A safety controller for a transition system $\Sigma_{\mathcal{P}} = (S,U_{A},F)$ and a specification $S^S_t\subseteq S, t = 0,1,2\ldots$ is a controller $C_t$ such that the following two properties hold
\begin{enumerate}
    \item $\dom(C_t) \subseteq S^S_t$ for all $t = 0,1,2 \ldots$ 
    \item for all $s\in\dom(C_t)$ for all $u\in C_t(s)$ the following inclusion is satisfied $F(s,u)\subseteq\dom(C_{t+1}).$
\end{enumerate}
\end{defn}
Here $\dom(C_t)=\{s\in S\mid C_t(s)\neq\emptyset\}$ is a domain of the controller and the following true; if $C_t$ is a safety controller, s.t. $\dom(C_t)\neq\emptyset$, then any controlled trajectory started at $\dom(C_0)$ never violates the safety requirements.
There are multiple ways to design such a controller $C_t$ for different tasks \cite{IVANOVA2021109, Pek}; however, this topic is out of the scope of this paper. We assume that a fallback safety controller is provided.

Let at the moment $t$ an agent be in a state $s_t$ and follow a policy $\pi$. We say that $a\in \enab_\pi(s_t)$ iff the agent may take an action $a\in A(s_t)$ (even with a probability of zero).

\begin{defn}
\label{verifiedsafe} A policy $\pi$ is verified safe at state $s_t$ (w.r.t a safety controller $C_t$) iff $F(s,a)\subseteq \dom(C_{t+1})$ for all $a\in\enab_\pi(s_t).$
\end{defn}




\subsection{Post-posed shields}
\label{classicalshield}
To provide hard safety guarantees for RL-based methods, one can design preemptive or post-posed shields that can be applied during the training or deploying phase~\cite{Alshiekh2018, krasowski2023provably}. This paper focuses on implementing post-posed shields for integration at deployment.

Let $Q^*(s, a)$ represent the optimal action-value function learned during training. The greedy policy $\pi^*(a|s)$ is the corresponding optimal policy for accomplishing the task. However, due to the sim-to-real gap or the absence of hard safety constraints during the training, deploying a robot with a policy $\pi^*(a|s)$ may lead to safety violations. In post-posed shielding, one monitors the RL agent behavior. Let at state $s_t \in \dom(C_t),$ and let an agent chose an action $a_t = a$ from the policy $\pi^*(a|s)$. If the policy is not verified safe (according to Definition \ref{verifiedsafe}); then the shield substitutes the action $a_t$ with a proven fallback strategy $u \in C_t(s_t)$, where $C_t$ is a safety controller for corresponding safety-relevant model $\Sigma = (S, U_A, F).$ Hence, when an RL agent is deployed together with a safety shield, it is guaranteed by design that only safe actions $a_t^{safe}(s,a)$ are applied to the system:
$$
 a_t^{safe}(s,a) = \left\{
\begin{aligned}
&a,\quad\mbox{ if } F(s,a) \subseteq \dom(C_{t+1})\\
&u\in C_t(s), \quad \mbox{ otherwise }
\end{aligned}\right.
$$
From Definitions \ref{MDP},\,\ref{SafetyRelevantAbstraction},\,\ref{SafetyController},\,\ref{verifiedsafe} it then immediately follows that system with a shield never violates safety constraints, as soon as $s_0\in\dom(C_0).$ 

A post-posed shield ensures safety but may degrade performance by forcing a safe but suboptimal policy in dangerous scenarios \cite{krasowski2023provably,Alshiekh2018}. Also, in most existing approaches \cite{Konighofer2020, Hunt2021, krasowski2023provably}, post-posed shields are designed with a minimum interference idea: the safety shield does not interrupt the agent until it is about to take an unsafe action. This might cause oscillations at the edge of the safe space, leading to system instability. Additionally, the robot might get stuck in a local minimum and never accomplish the task \cite{Hsu2024SF}. The next section proposes an approach to improve the post-posed shield design.

\section{Predictive safety shield based on Dyna-Q-learning}\label{sec:method}

\begin{algorithm}[t!]
\caption{SafetyFilter($\alpha, \gamma, \varepsilon, K, N$) \label{alg}}
\begin{algorithmic}[1]
\State {\bf Inputs}: Q-table $\mathbb{Q}$, Model $\mathbb{M}$, Initial State $s_0$
\State $t\leftarrow 0$
    \While{$s_t$ is not terminal}
\If{isVerifiedSafe(greedy\_policy($\mathbb{Q},s_t$), N) }
            \State $a\leftarrow$ greedy\_policy($\mathbb{Q}$,$s_t$)
        \Else
            \For {$s' \in \overline{\Reach}(s_t,N+1)$ \label{line4}}
                \State $\mathbb{Q}_W(s',a) \leftarrow \mathbb{Q}(s',a)$ for all $a\in A$ \label{line5}
            \EndFor
            \For{each of $K$ planning steps \label{LoopPlanningBegin}}
                \State (s', a') $\leftarrow$ RandSample($\mathbb{M}$, $\overline{\Reach}(s,N)$) \label{ProposedAction}
                \If{$\mathbb{M}(s',a_t^{safe}(s',a'))$ is not empty}
                    \State $(r', s'') \leftarrow$ from $\mathbb{M}(s',a_t^{safe}(s',a'))$ \label{ModelSample}
                    \State Update$(\mathbb{Q}_W, s',a',s'',\alpha,\gamma,r'$) \label{QValueModel}  \label{LoopPlanningEnd}
                \EndIf
            \EndFor
            \State $a \leftarrow \varepsilon$-greedy\_policy($\mathbb{Q}_W, s_t$) \label{EpsilonGreedy}
        \EndIf 
            \State $s_{t+1},r\leftarrow$ Move($s_t,a_t^{safe}(s_t,a)$) \label{Move}
            \State $\mathbb{M}(s_t,a_t^{safe}(s_t,a)) \leftarrow (r, s_{t+1})$ \label{ModelUpdate}
         \State $t\leftarrow t+1$
    \EndWhile   
\end{algorithmic}
\end{algorithm}

Contrary to many existing approaches, when the safety shield interrupts the agent only when it is about to take an unsafe action, we propose interrupting the agent if it takes an action that is verified unsafe in the next N steps. We then use a model of the world to simulate the two-player game, where the agent tries to maximize its reward, and the safety shield acts as a controversial agent. The robot then takes the most optimal action w.r.t. to the simulated game. We formalize this idea with Algorithm \ref{alg}.

Let us assume that a model $\mathbb{M}$ of the training environment is provided. It can be a distribution or sample model learned during training. It also can be a simulation engine if the agent is trained in a virtual world. We assume that for a given pair $(s, a),$ a model $\mathbb{M}(s,a)$ returns a next step $s'$ and a corresponding reward $r'$ or an empty tuple if a situation $(s, a)$ has not been modeled at all. We then initialize a Q-table $\mathbb{Q}$ with the optimal action-value function $Q^*(s, a)$ learned during training. At time step $t$, let an agent be at state $s_t$ and let it follow a deterministic greedy policy based on Q-table $\mathbb{Q}$. If this policy is verified safe for the next $N$ steps, we use an action generated with this policy to move to the next step. Otherwise, the generated action is replaced according to the procedure described in lines 7-14 of Algorithm \ref{alg}. We use the following function to verify that a policy $\pi$ is safe for the next $N$:\\
\noindent\hrule
\begin{algorithmic}
\State $i \gets t$, $S_0 \gets \{s_t\}$
\While{$i \leq t+N$} 
    \State $S_{i+1} \gets \bigcup_{s \in S_i} \bigcup_{a \in \enab_{\pi}(s)} F(s, a) $
    \If{$S_{i+1} \subseteq \dom(C_{i+1})$}
        \State $i \gets i + 1$
    \Else
        \State return false
    \EndIf
\EndWhile
\State return true
\end{algorithmic}
\noindent\hrule
\vspace{0.3cm}
If the policy was not verified safe, we first create a copy of a given $\mathbb{Q}$ function for all states that are reachable from state $s_t$ in no more then N+1 steps \ref{line4}-\ref{line5}. 
Then, in the loop \ref{LoopPlanningBegin}-\ref{LoopPlanningEnd}, we use a given model $\mathbb{M}$ to update the $Q_W(s, a)$ function while taking into account how the safety shield is going to act. Let us remark that an action proposed by the model at state $s'$ at line \ref{ProposedAction} might be unsafe. Since we are looking to get the most optimal path when only safe actions are performed, we then correct it with $a^{safe}(s', a')$. We then use $\varepsilon$-greedy policy (line \ref{EpsilonGreedy}) w.r.t, an updated version of the $Q_W$ function to move in the environment, while the safety shield ensures the taken action is safe (line \ref{Move}). 
 \begin{thm}
 Let the environment be modeled as MDP $\mathcal{P} = (S, A, T,r)$. Let a safety controller $C_t$ for a safety-relevant model $\Sigma_{\mathcal{P}} = (S,U_{A},F)$ and a specification $S^S_t\subseteq S, t = 0,1,2\ldots$ be provided. Then, an agent with a shield designed according to the Algorithm \ref{alg} never violates safety constraints as soon as the agent starts at $\dom(C_0)$.
 \proof The statement follows immediately from the fact that in line 15, only provably safe actions are taken.
 \end{thm}

Let us now discuss how the proposed safety shield affects the agent's performance. Let an agent move in the environment with a classical post-shield, described in Sec.~\ref{classicalshield}. Since the environment with the shield differs from the training environment, the agent will act suboptimal. One may propose to continue learning during the deployment phase to improve the agent's behavior. However, many resetting episodes may be needed before the agent gets the optimal state-value function $Q^*_{safe}(s,a)$. Instead, we propose to "simulate the learning process" in the robot's mind. We then claim that under the assumptions that the model $\mathbb{M}$ of MDP $\mathcal{P} = (S, A, T, r)$ is complete, number $N$ is large enough that $\overline{\Reach}(s_t, N+1)\subseteq \overline{\Reach}(s_t, N)$, and the deployment environment is static, the function $Q_W(s,a)$ converges to $Q^*_{safe}(s,a)$ as soon as set of parameters $\alpha,\gamma,\epsilon, K$ are appropriate \cite{Sutton1998}. Indeed, under such assumptions, Algorithm~\ref{alg} uses the model of the environment and safety shield to simulate the retraining process in the deployment environment. The assumption that refinement is static reflects the fact that function $a_t^{safe}(s, a)$ verifies safety w.r.t. constraints $S_t$ (even for the predicted states) and is unaware of potential environment changes in the future. Hence, we cannot fully model the interaction between the agent and the shield in the dynamic environment. 

We then propose to limit the prediction horizon over $N$ in future steps. Shorter horizons can lead to myopic, or short-sighted, behavior, where the controller prioritizes immediate improvements without considering future costs. This often leads to suboptimal performance, especially in tasks requiring strategic planning.
A longer horizon reduces myopia by giving the controller a "bigger picture" view, encouraging it to make decisions that might have lower immediate rewards but contribute to a higher long-term payoff. However, increasing $N$ increases the computation effort of the proposed algorithm. Moreover, commonly, we cannot speculate about safety in more than $N_{max}$ states due to engineering constraints, which gives us a natural upper bound on $N$.

Instead of creating and updating a local copy $\mathbb{Q}_W$, one can propose to update the original function $\mathbb{Q}$, both in the planning loop 9-13 and with a reward obtained in line 15. In such a case, Algorithm~\ref{alg} will correspond to a single run of Dyna-Q. Unfortunately, Dyna-Q is known for its poor performance when planning models include some positive errors \cite{Sutton1998}. That is precisely what happens when some unsafe move on the first steps of the run turns safe in the latter steps (see Example 2 for a more detailed explanation). Instead of updating the $\mathbb{Q}$ function, we provide robustness to environmental changes by leveraging the model's predictive control ideas: at every time step $t$ optimize behavior over horizon $N$, but then apply only the first action from the optimized sequence. Unfortunately, if $N$ is not big enough, this can cause agents to get stuck in an infinite loop and never complete the task. We can avoid these loops by enlarging the horizon, but there are some limits to increasing $N$. So, we propose to use the $\varepsilon$ greedy policy to escape the loops, at least eventually.

\section{Numerical Experiments}\label{sec:experiments}


For the experimental evaluation, we use the AI Safety Gridworlds \cite{leike2017aisafetygridworlds}. In our setup, the environment is a 6x6 or 7x7 grid world, where the agent occupies one cell. At every given time step, the agent is able to observe the coordinates of its position and choose an action from a predefined set $A$: moving up, down, left, right, and stop. Each action shifts the agent’s position by one cell in the corresponding direction. At the beginning of each episode, the agent will be reset to its initial position, denoted as $I$. 
The reward for each time step is: 
\begin{equation}
r_{t} = \alpha_{g} r_{g} + \alpha_{c} r_{c} + r_{d}
\end{equation}
where $\alpha_{g}$ and $\alpha_{c}$ are the coefficients indicating whether the target is reached and whether a collision occurs, respectively. The reward values for $r_{g}$, $r_{c}$, and $r_{d}$ can be found in Table~\ref{tab_hyperparameters}. 

First, we train our agent with a basic Dyna-Q learning algorithm with the hyperparameters shown in Table~\ref{tab_hyperparameters}. We choose a relatively high exploration rate to obtain a sufficiently well-modeled environment. 


We then consider the distributional shift problem: "How do we ensure that an agent behaves robustly when its test environment differs from the training environment? \cite{quinonero2022dataset}".
To investigate this, we evaluate different safety shield settings. The baselines are a Dyna-Q agent without a safety shield and with a baseline safety shield that does not consider performance criteria and just replaces an unsafe action with a given fallback strategy \cite{Alshiekh2018}. We selected two environments to demonstrate the capabilities of our approach, which we detail in the following.  

\begin{table}[htbp]
    \centering
    \caption{Summary of hyperparameters in the baseline Dyna-Q model}
    \begin{tabular}{|l|c|l|c|}
        \toprule
        Hyperparameter &  Value & Hyperparameter & Value\\ 
        \midrule
        Learning rate &  0.1 & Maximum episodes &  2000 \\       
        Discount factor  & 0.99 &  Goal-reaching reward ($r_{g}$)  & 50 \\
        Exploration rate  & 0.5 & Collision penalty ($r_{c}$) &  -50 \\
        Planning steps  & 10 &  Time-passage penalty ($r_{d}$)  & -1 \\
        Max steps per episode  & 1000 \\
        \bottomrule
    \end{tabular}
    \label{tab_hyperparameters}
\end{table}

\subsection{Experiment settings}



To test robustness under the distributional shifts, we provide the obstacle environments as shown in Figure~\ref{fig_exp1} and \ref{fig_exp2}. The agent must find a path from the initial state $I$ (blue cell) to the goal state $G$ (green cell) without colliding with obstacles. The policy is learned based on the training environment, but must also perform well in the unseen test environment. It is important to note that the agent is trained on one variant of the obstacle world, e.g., left environments in Figure~\ref{fig_exp1} and \ref{fig_exp2}. So even a small change in the environment, e.g., freeing one cell, is a major distribution shift.


\subsubsection{Path modification with obstacles}

\begin{figure}[htbp]
    \centering
    \begin{minipage}{0.14\textwidth}
        \centering
        \includegraphics[width=\textwidth]{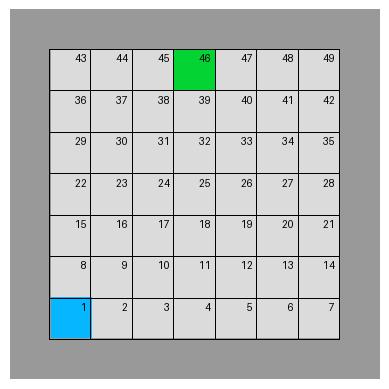}
    \end{minipage}
    \begin{minipage}{0.14\textwidth}
        \centering
        \includegraphics[width=\textwidth]{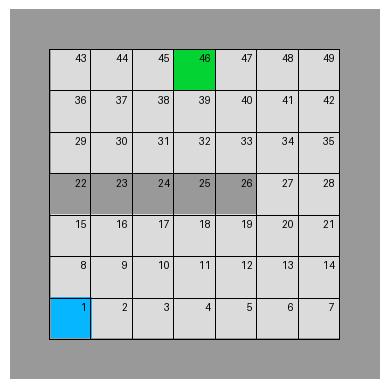}
    \end{minipage}
    \caption{Gridworlds for distributional shift problem. The left figure is the environment used for training, while the right one is for deployment.}
    \label{fig_exp1}
\end{figure}

As shown in Figure~\ref{fig_exp1}, we train in an environment where there are no obstacles blocking the midway, and in order to ensure determinism, we prioritize the movements with the same probability: up, down, left, right, and stop. In the test environment, we add obstacles in the grid cells 23-26, which intersect with the optimal path of the training environment.

\subsubsection{Target achievement in gating environment}

\begin{figure}[htbp]
    \centering
    \begin{minipage}{0.14\textwidth}
        \centering
        \includegraphics[width=\textwidth]{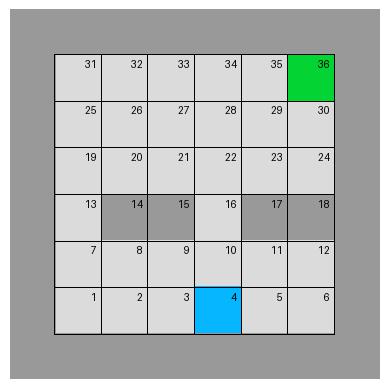}
    \end{minipage}
    \begin{minipage}{0.14\textwidth}
        \centering
        \includegraphics[width=\textwidth]{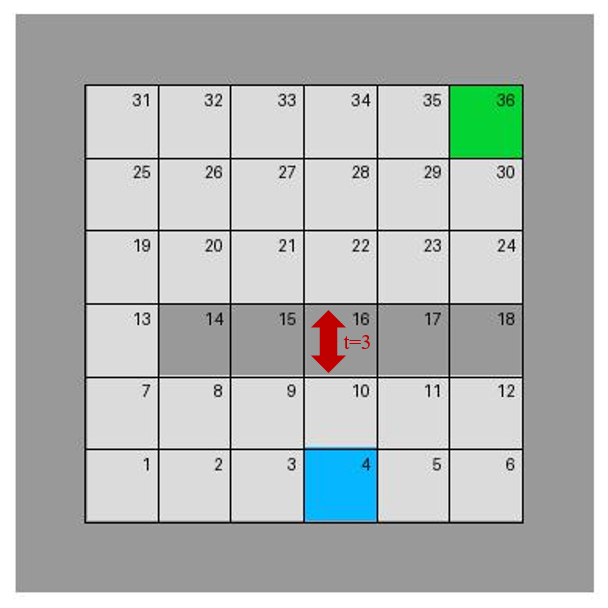}
    \end{minipage}
    \caption[A variant of classic grid maze]{A variant of classic grid maze.
    The gate in the left figure is opened, reflecting the training environment, while that in the right one is closed but will be opened at the third time step.}
    \label{fig_exp2}
\end{figure}

In order to evaluate the adaptability of the algorithms in changing environments, we designed classical gating scenarios, as shown in Figure~\ref{fig_exp2}. The rightmost grid cell $16$ is considered a gate, which remains open in the training environment but is closed in the testing environment until the third time step.

\begin{figure}[htbp]
    \centering
    \begin{minipage}{0.2\textwidth}
        \centering
        \includegraphics[width=\textwidth]{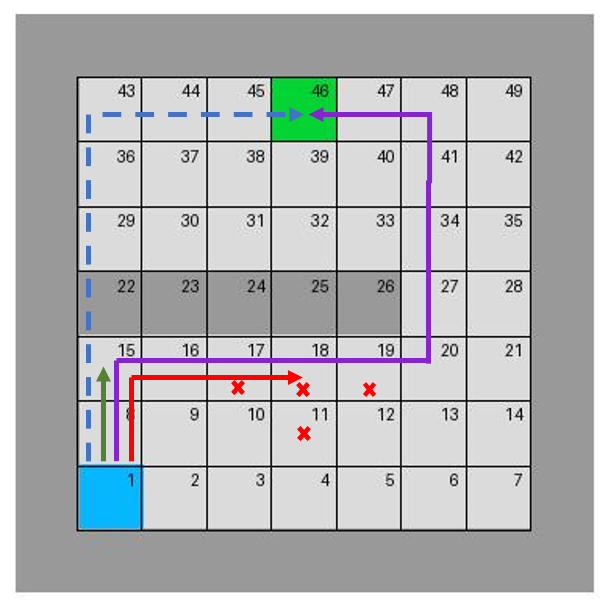}
    \end{minipage}
    \begin{minipage}{0.26\textwidth}
        \centering
        \includegraphics[width=\textwidth]{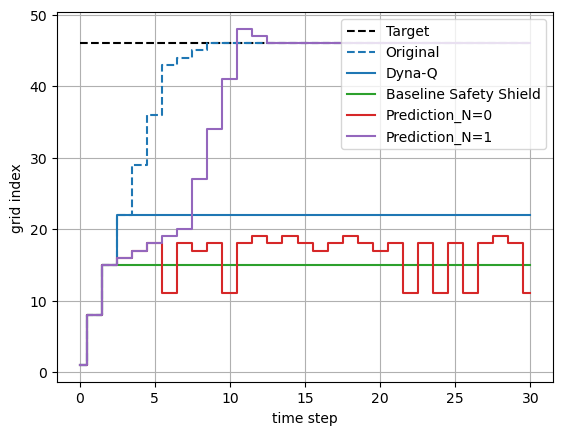}
    \end{minipage}
    \caption{Impact of prediction horizon on agent trajectories. The color of the trajectories on the left corresponds to that of the plots on the right, representing different test conditions. Red crosses mean that the agent is stuck in a loop in these grids.}
    \label{fig_exp1_result}
\end{figure}

\subsection{Results analysis}

\subsubsection{Path modification with obstacles}

We conducted several comparative experiments to explore the impact of different prediction horizons on the performance of the algorithm. The experimental results are shown in Figure~\ref{fig_exp1_result}. The blue path is the optimal path learned by the Dyna-Q agent in the training environment. However, with the appearance of new obstacles in the test environment, this path is now blocked. As a result, the Dyna-Q agent, without any safety measures, follows the blue path and ultimately collides with the newly introduced obstacle. 
Next, we investigate an agent equipped with a baseline safety shield~\cite{Alshiekh2018} that does not incorporate a prediction and replaces any risky action with a stop action. This approach prioritizes maximum safety by detecting obstacles and stopping at the last moment to avoid collisions. However, as indicated by the green path, while the agent remains safe, it eventually gets stuck. The red and purple line displays our algorithm, where the agent incorporates predictions for its decisions. We considered either a prediction horizon of zero or one depicted in red and purple, respectively. Within this local prediction window, our algorithm updates the Q-value (refer to Alg.~\ref{alg}). For $N=0$, the agent ultimately falls into a local optimum trap, where it is unable to reach the goal without significantly increasing randomness to explore other paths. As shown, even after 30 time steps, the agent remains stuck in a local loop indicated by the red crosses. However, if we set $N=1$, the agent discovers a new shortest route that leads to the target and also avoids collisions. According to the experimental results, our proposed algorithm, with a suitable predictive horizon, is indeed capable of addressing sudden environmental changes.

\subsubsection{Target achievement in gating environment}

\begin{figure}[htbp]
    \centering
    \begin{minipage}{0.2\textwidth}
        \centering
        \includegraphics[width=\textwidth]{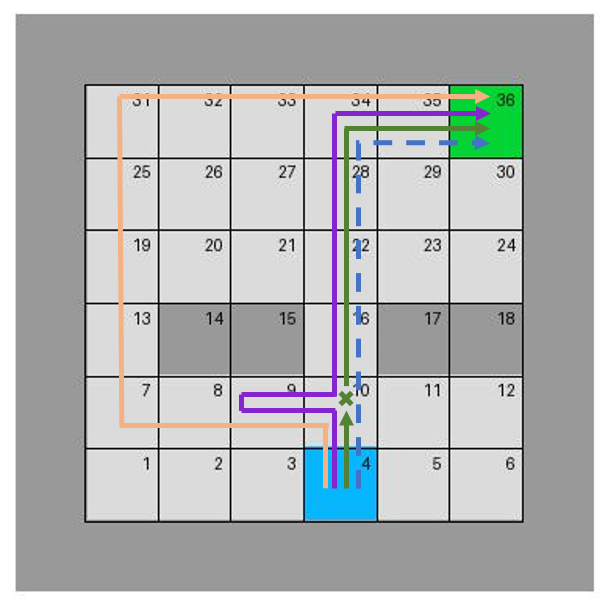}
    \end{minipage}
        \begin{minipage}{0.26\textwidth}
        \centering
        \includegraphics[width=\textwidth]{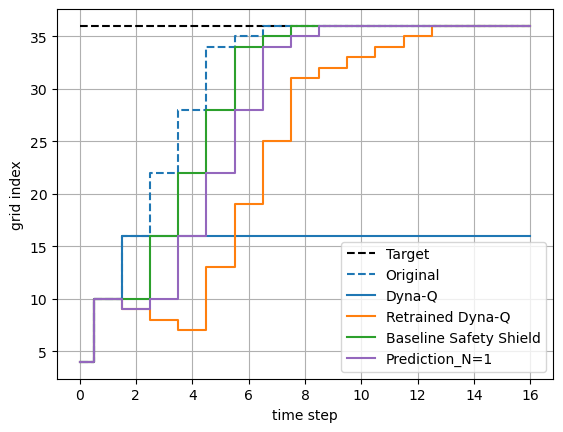}
    \end{minipage}
    \caption[]{Trajectory comparisons between retraining of RL algorithms such as Dyna-Q learning and our approach. The color of the trajectories on the left corresponds to that of the plots on the right, representing different test setups.}
    \label{fig_exp2_result}
\end{figure}

We compare the adaptability of the RL algorithm vanilla Dyna-Q with baseline safety shield (it replaces an unsafe action with a fallback stop strategy), and our method in the gating environment. The Dyna-Q model was trained in the environment on the left environment of Figure~\ref{fig_exp2}, while all tests were performed in the environment on the right environment of Figure~\ref{fig_exp2}. The experimental results are shown in Figure~\ref{fig_exp2_result}. The blue path represents the optimal route learned by the Dyna-Q agent in the training environment. In the test environment, the agent using Dyna-Q will still follow this path, resulting in a collision with the closed gate at time step 2. If we retrain this Dyna-Q agent in the test environment, we obtain the yellow path, which is not optimal since it takes 13 steps to reach the goal. The baseline safety shield agent waits at the door and enters immediately once the door opens, which incidentally leads it to obtain the optimal path. However, if the position or the opening time of the door were to change, the outcome could be significantly different. Next, we apply our algorithm with a predictive horizon $N=1$ and depict the result in purple. It is observed that, while the gate is closed, the agent maintains a path similar to that of the retrained Dyna-Q agent. However, at time step 3, our approach detects that the gate is open and subsequently reverts to the previous optimal path, passing through the gate to reach the goal. This allows our algorithm to reach the goal cell four steps earlier than the retrained agent. The experimental results demonstrate that our algorithm can respond in real time to environmental changes during execution, enabling the agent to adjust and plan an optimal path dynamically based on current conditions.



\begin{table}[h]
\centering
\caption{Path Optimality and Average Time Consumption}
\label{tab:method_comparison}
\begin{tabular}{ccccc}
\hline
    \multirow{2}{*}{\centering Method} & \multicolumn{2}{c}{\centering $\frac{\text{Time-steps}}{\text{Optimal Time-steps}}$} & \multicolumn{2}{c}{\centering $\frac{\text{Running Time}}{\text{Time-steps}}(s)$} \\
    \cline{2-5}
    & Env.1 & Env.2 & Env.1 & Env.2 \\
    \hline
    Dyna-Q                  & $\infty$  & $\infty$  & $\infty$  & $\infty$ \\
    Retrained Dyna-Q        & 1.000     & 1.354     & 0.718     & 0.729    \\
    Baseline Safety Shield  & $\infty$  & 1.000     & $\infty$  & 0.747    \\
    Prediction\_N=0         & $\infty$  & 1.073     & $\infty$  & 0.862    \\
    Prediction\_N=1         & 1.062     & 1.058     & 1.150     & 0.936    \\
    Prediction\_N=2         & 1.046     & 1.050     & 1.764     & 0.965    \\
    \hline
\end{tabular}
\end{table}


We also conducted a comparative analysis of path optimality and computation time across different methods. The ratio of actual time steps to optimal time steps indicates path efficiency, while the ratio of running time to time steps reveals computational cost per step, averaged across multiple initial states (i.e., cells 1-3,8-10 for the first environment and cells 3-5,9-11 for the second environment). 
For the first scenario, Table~\ref{tab:method_comparison} shows that if we retrain Dyna-Q in the environment with the obstacle, it learns the optimal path. However, it is important to note that this retraining process is resource-intensive. Our approach may slightly deviate from the optimal path due to the use of an epsilon-greedy policy. This policy occasionally results in the agent randomly selecting safe actions, which can lead to suboptimal paths. If we set the epsilon parameter to zero and choose a sufficiently large time horizon, the agent will follow the shortest path to the goal. Nevertheless, we prefer to keep epsilon greater than zero to avoid infinite loops, even for shorter time horizons. It is also worth noting that the baseline safety shield — which simply replaces an unsafe action with a stop action — is stuck indefinitely in this scenario. For the second scenario, our approach outperforms retrained Dyna-Q since it can adapt to the change of the environment faster. In this particular scenario, baseline safety shield outperforms our approach. However, if the opening time of the door were to change, the outcome could be significantly different. For example, if the door remains closed until time step 7 or later, one should take the path around the obstacle. Our algorithm can adapt to this situation, but the baseline safety shield would cause the agent to remain stuck next to the door until it opens. In summary, our safety shield can adapt to environmental changes on the fly, and a shield with longer prediction horizons is generally more optimal. However, the computational cost per step increases with a longer horizon.





\section{Discussion}\label{sec:discussion}

Our experimental setup demonstrates the advantages of the proposed approach compared to existing baselines. While we observe a performant path in the dynamic gate environment, our approach does not ensure optimality for dynamic obstacles that are moving in the environment. For that, we would need to extend the safety-relevant model so that it predicts the possible future occupancies of dynamic obstacles. To ensure hard safety guarantees, we would need to use an overapproximating prediction. For example, let us assume a dynamic obstacle can at most move one cell. For the third prediction step, the potential occupancy would be at most 25 cells, which is more than half of all grid cells of our environments. This illustrates that performant and safe policies are especially difficult to find in highly dynamic environments and thus an active field of research.

Most real-world tasks are solved in continuous space. One common approach to making algorithms operating in discrete domains applicable is to discretize the state and action space. However, this can render the algorithms infeasible due to a state space explosion when fine discretization is required. To address this, one could use automata learning to obtain a mapping between the discrete and the continuous space that is only refined where necessary \cite{Tappler2022}. An alternative approach is learning in the continuous space directly. For that, a parallel work \cite{banerjee2024dynamic} proposed a multi-step safety shield that leverages Monte Carlo tree search to determine an optimal and safe plan. However, for long-time horizons and high-dimensional state and action spaces, it remains unclear if the proposed approach would be computationally feasible. This is mainly relevant when considering dynamic obstacles, which freely move in the space. Thus, an extension to continuous space would additionally require different methods for predicting the movements of dynamic obstacles, e.g., reachability analysis \cite{althoff2021set}.


\section{Conclusion}\label{sec:conclusion}
In this work, we propose a predictive safety shield for discrete spaces for the model-based RL algorithm Dyna-Q. Importantly, we derive that our local Q-function update achieves optimality given that the full environment can be observed and only static obstacles are present. In our experiments, we validate our approach and investigate the influence on optimality for different prediction horizons. In particular, we demonstrate that in simple dynamic environments, our approach obtains close to or optimal solutions without additional training. Future work could extend our approach to make it applicable to highly dynamic environments in continuous space so that it becomes feasible to shield autonomous systems in the real world.

\bibliographystyle{IEEEtran}
\bibliography{ref}


\end{document}